\documentclass[sigconf,screen]{acmart} 
\AtBeginDocument{%
  }

\setcopyright{acmlicensed}
\copyrightyear{2018}
\acmYear{2018}
\acmDOI{XXXXXXX.XXXXXXX}
\acmConference[MM '26]{The 34th ACM International Conference on Multimedia}{November 10--14, 2026}{Rio de Janeiro, Brazil}
\acmISBN{978-1-4503-XXXX-X/2018/06}

\def\eg{\emph{e.g.}} 
\def\ie{\emph{i.e.}}

\usepackage[table]{xcolor}
\definecolor{oursblue}{RGB}{228,237,248}
\acmSubmissionID{1787}

\begin{document}

\title{CogPortrait: Fine-Grained Eye-Region Control in Portrait Animation via Hierarchical Agent Planning}

\author{He Feng}
\email{fenghe021209@gmail.com}
\affiliation{%
  \institution{Harbin Institute of Technology}
  \city{Harbin}
  \country{China}
}

\author{Yongjia Ma}
\authornote{Project Leader.} 
\email{maguire9993@gmail.com}
\affiliation{%
  \institution{Li Auto}
  \city{Beijing}
  \country{China}
}

\author{Donglin Di}
\email{didonglin@lixiang.com}
\affiliation{%
  \institution{Li Auto}
  \city{Beijing}
  \country{China}
}

\author{Lei Fan$\dagger$}
\email{lei.fan1@unsw.edu.au}
\affiliation{%
  \institution{University of New South Wales}
  \city{Sydney}
  \country{Australia}
}  

\author{Tonghua Su$\dagger$}
\email{thsu@hit.edu.cn}
\affiliation{%
  \institution{Harbin Institute of Technology}
  \city{Harbin}
  \country{China}
}

\renewcommand{\shortauthors}{He Feng, Yongjia Ma, Donglin Di,  Lei Fan$\dagger$, \& Tonghua Su$\dagger$}

\renewcommand{\shortauthors}{He Feng, Yongjia Ma, Donglin Di,  Lei Fan, \& Tonghua Su}


\begin{abstract}
Portrait animation methods have achieved substantial visual quality and lip synchronization, but fine-grained manipulation of the eye region still faces a trade-off between input granularity and motion accuracy.
Existing methods using emotion labels or coarse text prompts are insufficient for describing subtle ocular dynamics, whereas approaches based on Action Units or driving videos provide higher fidelity at the cost of a heavier input burden.
These limitations are still restrictive for beyond-emotion states (\eg, thinking) and drowsiness.
In light of the above, we propose CogPortrait, a two-stage framework
that generates portrait animations from high-level labels.
In the first stage, three chain-of-thought Multimodal Large Language Models (MLLMs) agents compile high-level labels into facial keypoints through temporal event planning, prototype retrieval, and composition from a real-behavior library, and semantic-physiological constraint enforcement.
In the second stage, a DiT-based video generation backbone synthesizes the final animation conditioned on the keypoints, reference portrait, audio, and text prompt, enhanced by a dynamic classifier-free guidance strategy with eye-region-aware reweighting and KTO-based refinement for boundary cases. 
We further introduce the EMH benchmark covering diverse emotions and 
beyond-emotion categories with two AU-level metrics for evaluating 
fine-grained eye-region and head-motion control. 
Extensive experiments on HDTF and the EMH benchmark demonstrate that CogPortrait achieves more precise eye-region control than existing methods while maintaining superior visual quality and identity consistency.   


\end{abstract}




\renewcommand\footnotetextcopyrightpermission[1]{}
\settopmatter{printacmref=false} 

\maketitle

\begingroup
\renewcommand{\thefootnote}{}
\footnotetext{$^{\dagger}$Lei Fan and Tonghua Su are the corresponding authors.}
\endgroup

\section{Introduction}
\label{sec:intro}
Portrait animation \cite{zhang2023sadtalker, cui2024hallo3, feng2025ditalker} aims to synthesize photorealistic face videos from a single reference image, conditioned on multimodal signals such as audio \cite{hong2025audio, zhai2023talking} and facial keypoints \cite{feng2024one, hu2024animate}.
With broad applications in virtual reality \cite{lee2025audio}, human-computer interaction \cite{dollinger2023embodied}, and film production \cite{alexander2010digital}, this field has attracted rapidly growing research interest \cite{cui2024hallo2,yang2024megactorsigmaunlockingflexiblemixedmodal,drobyshev2022megaportraits}.
By adopting diffusion models \cite{yang2024cogvideox, wan2025wan, xu2024easyanimate} as the generation backbone, current methods \cite{stypulkowski2024diffused, feng2025ditalker,chen2024echomimic} have achieved substantial improvement in visual quality \cite{cui2024hallo2}, lip synchronization \cite{ji2025sonic}, and temporal stability \cite{ki2025float}.
Building on these advances, recent efforts \cite{zhou2025gohd,ma2025facetimeline,gururani2023space} have begun to pursue finer control over the eye-region and head pose, as these cues not only affect the perceived naturalness and vividness of animated results \cite{garau2003avatar,canales2023gaze}, but also convey beyond-emotion states such as cognitive effort and drowsiness, thereby expanding the expressive range of portrait animation.

\begin{figure*}[!t]
    \centering
\includegraphics[width=0.99\linewidth]{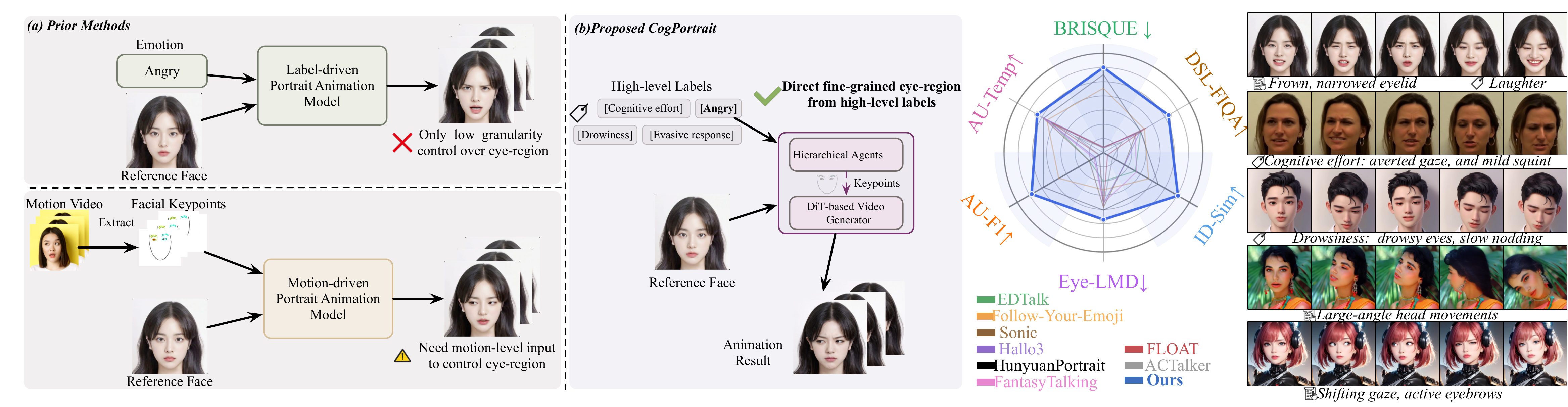}
    \caption{
    Comparison with prior portrait animation paradigms, quantitative results on the EMH benchmark, and OOD cases generated by CogPortrait.
    Left: comparison of prior label-driven  \cite{tan2025edtalk,ki2025float} and motion-driven methods \cite{guo2024liveportrait,xu2025hunyuanportrait} with the proposed CogPortrait framework, highlighting direct fine-grained eye-region control from high-level labels.
    Central: radar-chart comparison with baselines \cite{tan2025edtalk,ma2024follow,ji2025sonic,cui2024hallo3,xu2025hunyuanportrait,wang2025fantasytalking,ki2025float,hong2025audio}
    on the EMH benchmark, where larger areas indicate better performance. Right: OOD cases generated by CogPortrait, including fine-grained instruction, laughter, cognitive effort, drowsiness, large-angle head movement, and control of gaze and eyebrow motion.}
    \label{fig:1}
\end{figure*}

Existing approaches \cite{xu2025hunyuanportrait, ma2024follow, zhou2025gohd} capable of controlling the eye region and head pose can be broadly divided into two categories. The first category is driven by high-level labels or coarse text prompts \cite{tan2025edtalk,wang2024instructavatar,liu2025moee,ki2025float,cui2024hallo3}. These methods accept simpler inputs but usually operate only at the level of coarse semantic categories (\textit{e.g.,} \textit{``anger''}) or macro motion directives (\textit{e.g.,} \textit{``turn your head to the left.''}), making them insufficient for fine-grained eye-region dynamics, such as asymmetric blinks or gaze aversion.
The second category is driven by motion-level signals such as facial keypoints \cite{feng2024one, ma2024follow, gururani2023space}, Action Units \cite{lyu2026auhead, ma2025facetimeline}, and driving videos \cite{xu2025hunyuanportrait,guo2024liveportrait, hong2025audio}. These methods provide finer control but require the user to supply or manually construct motion-level driving signals, which increases input burden and limits flexibility.
This limitation is more pronounced for irregular patterns such as asymmetric eyebrow motion and large-angle head movements, where suitable driving signals are difficult to obtain, and the patterns are underrepresented in training data \cite{zhang2021flow,yu2023celebv}, leading to degraded control accuracy and identity consistency. Overall, directly achieving fine-grained control over eye-region dynamics from high-level labels remains an open challenge.

In light of the above, we propose \textbf{CogPortrait}, a two-stage framework for fine-grained eye-region control in portrait animations from high-level labels (see Figure~\ref{fig:1}).
We define the target beyond-emotion states through coordinated changes in gaze, eyelid, blink, eyebrow, and head pose (\eg, cognitive effort via coupled gaze-head shifts; drowsiness via prolonged blinks and droopy eyelids).
In the first stage, we compile high-level labels into executable facial keypoints. Rather than directly generating motion signals from the textual \cite{hang2023lang} or category conditions \cite{wang2024instructavatar}, which we observe tends to produce over-smoothed trajectories that miss the abrupt, irregular dynamics of real facial behavior, we formulate this stage as a structured control-generation process using three chain-of-thought (CoT) \cite{wei2022chain} reasoning MLLM agents: a \textit{planning agent} that decomposes the target label into temporally structured events, a \textit{composition agent} that retrieves and assembles behavioral prototypes from a curated library to preserve genuine temporal dynamics, and a \textit{critic agent} that enforces semantic consistency and physiological 
plausibility.



In the second stage, we employ a DiT-based video diffusion backbone \cite{wan2025wan} conditioned on the reference portrait, driving audio, text prompt, and the generated facial keypoints to produce the final animation. To further improve generation quality, we address two orthogonal issues in this stage.
First, a single global classifier-free guidance (CFG) scale tends to weaken high-frequency eye-region motion~\cite{jin2026stagewise} and introduce global color shift~\cite{sadat2024eliminating} when the reference image has a uniform background; we address this with a dynamic CFG strategy featuring eye-region-aware reweighting in both the temporal and spatial dimensions.
Second, the model often struggles with boundary cases such as asymmetric eyebrow control and identity degradation under large-angle head movements due to their underrepresentation in training data; we address this with Kahneman-Tversky Optimization (KTO)-based~\cite{ethayarajh2024kto} refinement using curated desirable and undesirable samples for these long-tail patterns.

To evaluate fine-grained eye-region and head-motion control, we build the \textbf{EMH benchmark} covering six core emotions from MEAD \cite{wang2020mead} and six beyond-emotion states (\eg, cognitive effort, drowsiness, social engagement), and introduce two AU-level metrics: AU-F1 for activation correctness and AU-Temp for temporal trajectory fidelity.
As shown in Figure~\ref{fig:1}, CogPortrait achieves state-of-the-art performance on the EMH benchmark and generates faithful fine-grained eye-region dynamics across diverse beyond-emotion states.

Our contributions can be summarized as follows:
\begin{itemize}
\item We present CogPortrait, a two-stage framework that compiles high-level labels into fine-grained eye-region facial keypoints via structured agent-based planning, prototype retrieval, and constraint enforcement, and generates portrait animation conditioned on these keypoints.
\item We introduce a dynamic CFG with an eye-region-aware reweighting strategy and KTO-based refinement for faithful rendering of fine-grained eye-region motions and irregular motion patterns.
\item We propose the EMH benchmark with two AU-level metrics for evaluating fine-grained eye-region and head-motion patterns.
\end{itemize}



\begin{figure*}[!t]
    \centering
\includegraphics[width=0.999\linewidth]{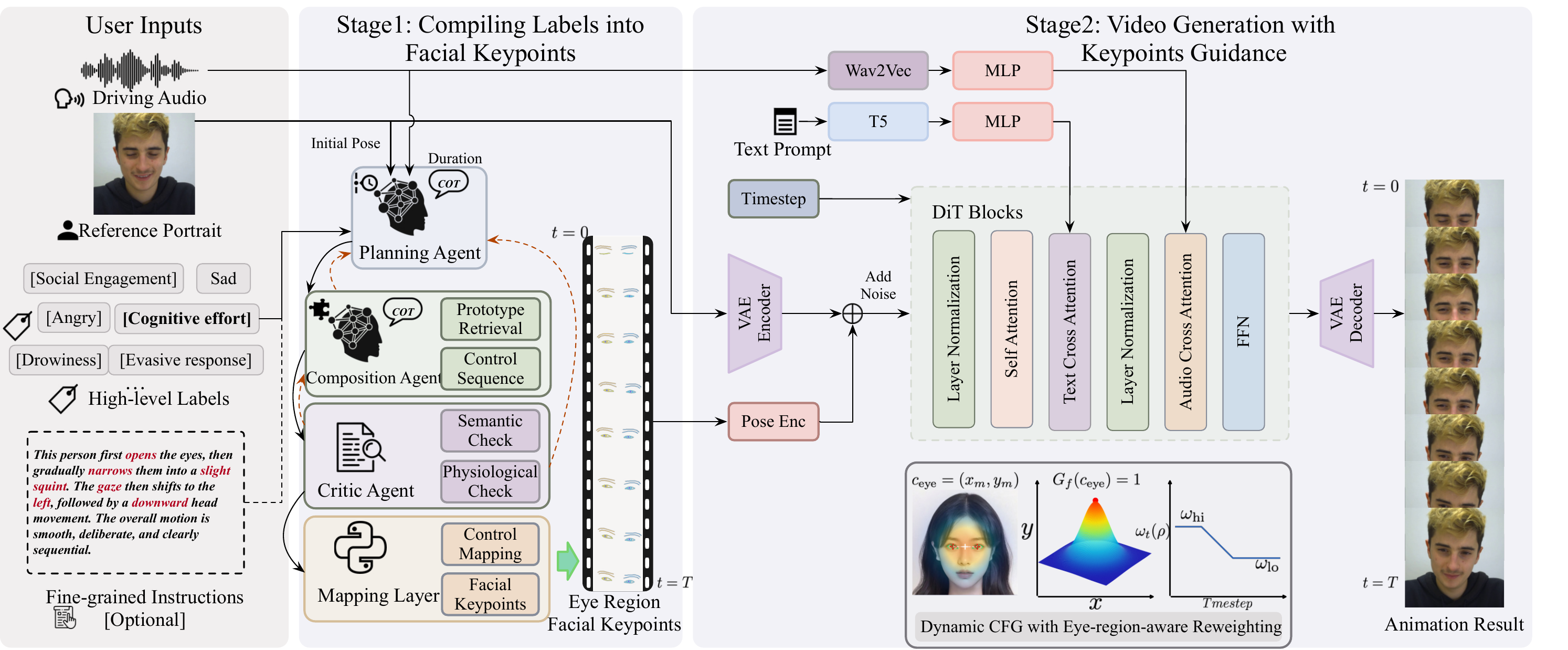}
\caption{Overview of the proposed CogPortrait. Stage~1 compiles a high-level label, a reference portrait, and optional fine-grained instructions into facial keypoints through four layers: Event Planning produces staged events, a CoT composition agent performs prototype retrieval to construct a control sequence, a critic agent applies semantic and physiological checks to refine the sequence, and a mapping layer projects the refined controls into facial keypoints. The critic can send revision signals back to the composition or planning layer. Stage~2 generates the final portrait animation from the reference portrait, driving audio, a text prompt, and the facial keypoints via a DiT-based generation backbone with dynamic CFG with eye-region-aware reweighting.}
    \label{fig:pipeline}
\end{figure*}

\section{Related Work}
\subsection{Portrait Animation Methods}
Recent portrait animation methods based on diffusion \cite{yang2024megactor,wei2024aniportrait} or DiT backbones \cite{yang2024megactorsigmaunlockingflexiblemixedmodal} have achieved superior visual quality, lip synchronization, and temporal stability~\cite{tian2025emo,liu2025moee,cui2024hallo2,chen2024echomimic,cui2024hallo3}. Several methods further enable control over expression and affective states through emotion-labeled audio~\cite{ki2025float,ji2025sonic}, text instructions~\cite{wang2024instructavatar,liu2025moee}, or temporal Action Unit sequences~\cite{lyu2026auhead}, but they typically operate only at the level of coarse semantic categories or macro motion directives, and cannot directly specify fine-grained eye-region dynamics such as gaze, blink, and eyebrow motions.

Another line of works \cite{zhou2025gohd,ma2025facetimeline} leverages explicit motion signals, such as driving videos~\cite{xu2025hunyuanportrait,guo2024liveportrait,rochow2024fsrt} and expression-aware landmarks~\cite{ma2024follow,hu2024animate}, to achieve fine-grained control over the eye-region and head pose.       
More closely related to ours, GoHD~\cite{zhou2025gohd} enables explicit gaze control through gaze-oriented motion factors, and Timeline Control~\cite{ma2025facetimeline} provides fine-grained temporal control over specific facial regions via user-defined multi-track timelines.   Although these methods offer finer control over the eye region and head pose, they depend on motion-level driving signals as input and remain unreliable for irregular patterns, whereas our framework directly generates fine-grained eye-region keypoints from either high-level semantic labels or fine-grained instructions (\eg, ``raise your left eyebrow'').


\subsection{Agents for Video and Motion Generation} 
Agent-based frameworks that leverage large language models and chain-of-thought reasoning have recently demonstrated strong capabilities for high-level planning in video  generation~\cite{wu2025automated,xu2025filmagent} and motion synthesis. VideoAgent~\cite{soni2025videoagent} and AniME~\cite{zhang2025anime} employ multi-agent collaboration for automated video production, while             
Motion-Agent~\cite{wu2024motion} and L2-EMG~\cite{wang2026towards} support the generation and continual refinement of high-level motion behaviors across open scenarios.
Stage 1 of our framework also employs MLLM agents and chain-of-thought reasoning to compile high-level semantic labels into facial keypoints through temporal event planning, prototype retrieval, composition, and constraint enforcement.



\section{Methodology}
Our framework generates portrait animation that faithfully renders specified eye-region and head-motion patterns from high-level labels, a reference portrait, driving audio, and optional fine-grained instructions (\eg, a target gaze direction, a head-motion trajectory, or a desired eyebrow amplitude). 

An overview of our framework is illustrated in Figure~\ref{fig:pipeline}.
Stage~1 compiles high-level labels, a reference portrait, and optional fine-grained instructions into eye-region facial keypoints, with the temporal length determined by the driving audio duration (Sec.~\ref{sec:stage1}).
Stage~2 generates the final portrait animation from the reference portrait, driving audio,
a text prompt, and facial keypoints (Sec.~\ref{sec:stage2}).
The prototype library and the EMH benchmark are described in Sec.~\ref{sec:prototype}.

\subsection{Preliminaries}
\label{sec:preliminaries}
\noindent\textbf{Flow Matching.}
Flow Matching (FM) models \cite{lipman2023flow} generate data by learning a time-dependent velocity field that transports a Gaussian prior distribution to the target data distribution in latent space. Given a video $X \in \mathbb{R}^{T \times H \times W \times 3}$, where $T$ denotes the number of frames, $H$ and $W$ denote the frame height and width, and $3$ denotes the RGB channels, a pretrained 3D VAE encoder $\mathcal{E}$ \cite{kingma2013auto} compresses $X$ into a latent representation $x_1 = \mathcal{E}(X) \in \mathbb{R}^{T' \times H' \times W' \times C}$. Here, $T' = T/\tau$ is the compressed temporal length with temporal compression factor $\tau$, $H' = H/s$ and $W' = W/s$ are the compressed spatial dimensions with spatial compression factor $s$, and $C$ denotes the latent channel dimension. During training, Gaussian noise $x_0 \sim \mathcal{N}(0, I)$ is sampled, and a velocity prediction network $v_\theta(\cdot)$ is trained to predict the transport velocity from an interpolated latent $x_t$ at timestep $t \in [0,1]$ under condition $c$ (\eg, audio or keypoints). Its objective is
\begin{equation}
\mathcal{L}(\theta) = \mathbb{E}_{x_1, c,\, t \sim \mathrm{Unif}[0,1],\, x_0 \sim \mathcal{N}(0, I)} \left[ \left\| v_\theta(x_t, t, c) - (x_1 - x_0) \right\|^2 \right].
\end{equation}
During inference, the learned velocity field is integrated from a Gaussian sample to produce the generated latent $\hat{x}_1$, which is then decoded by a VAE decoder $\mathcal{D}$ into the final video $\hat{X}$. We adopt Wan2.2~\cite{wan2025wan}, a flow-matching-based video diffusion model, as the generation backbone.

\noindent\textbf{Kahneman-Tversky Optimization.}
Kahneman-Tversky Optimization (KTO)~\cite{ethayarajh2024kto} aligns a model with human preferences using only per-sample binary feedback (desirable or undesirable), without requiring paired preference data. The implicit reward is defined as $r_\theta(c, x) = \log \frac{\pi_\theta(x|c)}{\pi_{\mathrm{ref}}(x|c)}$, where $\pi_\theta$ is the learned policy and $\pi_{\mathrm{ref}}$ is a frozen reference policy. The training objective is formulated as:
\begin{equation}
\mathcal{L}_{\mathrm{KTO}} = \mathbb{E}_{c, x} \left[ w(y) \cdot \left(1 - \sigma\left( \beta \cdot r_\theta(c, x) - z_{\mathrm{ref}} \right)\right) \right],
\end{equation}
where $z_{\mathrm{ref}}$ is a KL-based reference point estimated from the dataset, $w(y)$ assigns asymmetric weights to desirable and undesirable samples, $\sigma(\cdot)$ is the sigmoid function, and $\beta$ is the deviation parameter.

\begin{figure}[!t]
\centering\includegraphics[width=0.999\linewidth]{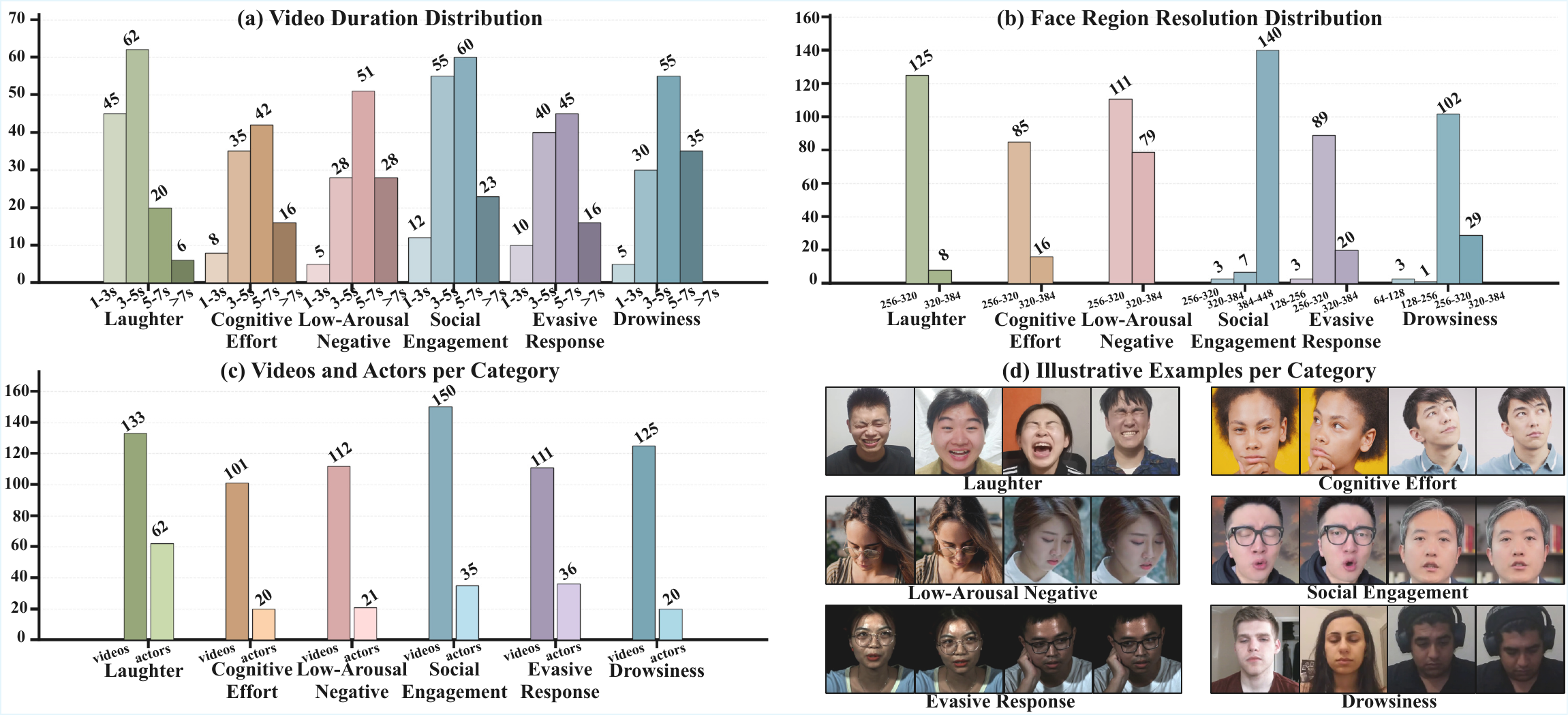}
    \caption{Overview of the EMH benchmark. The figure shows the video-duration distribution, face-region resolution distribution, numbers of videos and actors, and representative sample cases across the beyond-emotion states.}
    \label{fig:benchmark}
\end{figure}

\subsection{Prototype Library and the EMH Benchmark}
\label{sec:prototype}
\noindent\textbf{Prototype Library.}   When MLLM agents directly generate fine-grained controls, they often produce over-smoothed AU trajectories, which lead to suboptimal portrait animation results.
To address this issue, we construct a prototype library $\mathcal{L}$ for retrieval in Stage~1.
Each prototype is represented as $p=\{y,\{c_t\}_{t=1}^{T},\{k_t\}_{t=1}^{T}\}$, where $y$
denotes the category label, $c_t=[a_t,g_t,r_t]$ denotes the control state at frame $t$, and $k_t$ denotes the corresponding facial keypoints.
The control state comprises 17 channels, including 10 AU channels for eyebrow, eyelid, and blink, 4 gaze channels for gaze direction, and 3 head-pose channels.
Given real videos annotated with AU intensities and matched neutral references, we extract aligned keypoints $\tilde{k}_t$ and head pose.
Since each AU channel corresponds to a specific facial muscle action with well-defined physical meaning, these prototypes preserve genuine temporal dynamics of real facial behavior rather than arbitrary synthesized trajectories.
We then define an identity-specific neutral baseline $b$ as the mean keypoint under rest expression, and compute the residual deformation $d_t=\tilde{k}_t-b$. To reduce the coupling among gaze, eyelid, and head motion, we recover non-gaze controls $q_t=[a_t,r_t]$ and gaze controls $g_t$ separately from the residuals. The recovered controls and keypoint trajectories are then organized into structured prototypes for retrieval in Stage~1 and animation in Stage~2.
Additional construction details are provided in the \textit{supp.}.

\noindent\textbf{EMH Benchmark.}
Existing portrait animation benchmarks \cite{wang2020mead,yu2023celebv,di2024facevid} focus on basic emotions and lack beyond-emotion states whose primary visual cues are eye-region dynamics rather than overt expressions; we therefore introduce the EMH benchmark to evaluate fine-grained eye-region and head-motion pattern generation from high-level labels.
As shown in Figure~\ref{fig:benchmark}, the benchmark contains 12 categories in total, including six core-emotion categories and six beyond-emotion states, and summarizes their video-duration distribution, face-region resolution distribution, numbers of videos and actors, and representative sample cases. The six core-emotion categories, \emph{sadness}, \emph{fear}, \emph{disgust}, \emph{contempt}, \emph{anger}, and \emph{surprise}, are curated from MEAD~\cite{wang2020mead} and contain 120, 122, 115, 121, 120, and 124 videos from 33, 38, 30, 33, 32, and 34 actors, respectively. The six beyond-emotion categories include \emph{laughter}, curated from DH-FaceLolVid~\cite{liu2025moee}, with 133 videos from 62 actors; \emph{cognitive effort}, curated from UBFC-Phys~\cite{9346017}, with 101 videos from 20 actors; \emph{low-arousal negative state}, curated from EAV~\cite{lee2024eav}, with 112 videos from 21 actors; \emph{social engagement}, curated from DH-FaceVid-1K~\cite{di2024facevid}, with 150 videos from 35 actors; \emph{evasive response}, curated from SEUMLD~\cite{xu2025multimodal}, with 111 videos from 36 actors; and \emph{drowsiness}, curated from UTA-RLDD~\cite{ghoddoosian2019realistic}, with 125 videos from 20 actors. Each case is paired with a text prompt and manual eye-region dynamic annotations; for example, a drowsiness clip is annotated as: frames 0--a exhibit droopy eyelids, frames a--b contain a prolonged blink, and frames b--end show a drowsy head nod.  
For this benchmark, we release only frame indices and prompt annotations, while access to the original videos remains subject to the licenses and usage terms of the source datasets.
Additional details are provided in \textit{supp}.

\subsection{Stage 1: Compiling Labels into Keypoints}
\label{sec:stage1}
Stage~1 compiles a target label $y$, a reference portrait $I_{\mathrm{ref}}$, the driving audio duration, and  optional fine-grained instructions $u$ into a keypoint sequence $K_{1:T}=\{k_t\}_{t=1}^{T}$, where $k_t$ denotes the facial keypoints at frame $t$. The main difficulty is that the same label admits multiple plausible temporal realizations.
To address this ambiguity, we adopt a hierarchical agent-based strategy consisting of two CoT reasoning MLLM agents and a critic agent: the planning agent decomposes the label into temporally structured events, the composition agent retrieves and composes prototypes into an initial control sequence, and the critic agent enforces semantic consistency and    
physiological plausibility.  

\noindent\textbf{Temporal Event Planning.}
We use the \textit{cognitive effort} as an example throughout Stage~1 and provide the agent prompts and category-to-control mappings for all other categories in \textit{supp}.
The planning agent first decomposes the target label and optional fine-grained instructions into a sequence of staged events, denoted by $E=\{e_k\}_{k=1}^{N_e}$, whose total duration satisfies $\sum_{k=1}^{N_e}\Delta t_k=T$. Each event specifies its timestamp range, frame span, stage semantics, and local constraints, converting the high-level label into a set of temporally structured and retrievable local behavioral units.
A representative prompt fragment is
``\textit{You are a psychology expert. Given a target state label, optional fine-grained instructions, and the target duration, decompose the behavior into a short sequence of staged facial events. Output the result in a structured format, and for each event, provide its timestamp range, frame span, and key gaze, eyelid, blink, eyebrow, and head-pose changes.}''
For example, given the input \textit{cognitive effort} with the instruction \textit{look slightly to the left, then lower the head smoothly}, the planning agent may output ``\textit{[0.00s--0.24s, 1--6f] attentive onset; [0.24s--0.72s, 7--18f] gaze shifts left with a mild squint; [0.72s--2.00s, 19--50f] head lowers slightly while maintaining the squint}.''

\noindent\textbf{Prototype Retrieval and Composition.}
The composition agent takes the staged event sequence $E$, the reference portrait $I_{\mathrm{ref}}$, and the
optional fine-grained instructions $u$ as input, and constructs an initial frame-wise control sequence      
$C_{1:T}^{(0)}=\{c_t^{(0)}\}_{t=1}^{T}$.
To avoid an abrupt pose jump at the first frame, the initial pose estimated from the reference portrait is provided as an additional input.
For each staged event, the agent converts the event description into target channel requirements over the predefined 17-channel control space, and retrieves prototypes from the library $\mathcal{L}$ (Sec.~\ref{sec:prototype}) by minimizing a channel-wise
matching distance:
  \begin{equation}
  D(q, p) = \sum_{j=1}^{17} w_j \, |q_j - \bar{p}_j|,  
  \end{equation}                                       where $q_j$ denotes the target value on channel $j$, $\bar{p}_j$ denotes the mean value of prototype $p$ on channel $j$, and $w_j$ are channel-type weights with AU channels weighted more heavily than gaze and head-pose channels.
The retrieved prototypes are then trimmed to the required duration, rescaled in amplitude to match the
target ranges, and stitched across events into $C_{1:T}^{(0)}$.
A representative prompt fragment is            
``\textit{You are an expert in facial behavior analysis. Given a staged event sequence from the planning agent, a reference portrait, and optional fine-grained instructions, retrieve and compose structured control priors for each stage. Output the result in a structured format, and for each stage, provide the valid AU, gaze, and head-pose ranges, duration, and required edits for composition.}'' For example, given the staged events of \textit{cognitive effort}, the composition agent may output ``\textit{t1: AU5 in [0.15, 0.35], direct gaze,
stable head pose; t2: AU4 in [0.10, 0.20], AU7 in [0.12, 0.25], gaze\_left in [0.15, 0.30], gaze\_up in [0.10, 0.25]; t3: AU4 in [0.05, 0.12], AU7 in [0.08, 0.18], gaze\_left in [0.25, 0.50], pitch in
$[-12^\circ,-5^\circ]$}.''

\noindent\textbf{Constraint Enforcement.}
The critic agent checks and refines the initial control sequence before it is mapped to keypoints. Its input includes the target label $y$, the staged event sequence $E$, the optional fine-grained instructions $u$, and the initial control sequence $C_{1:T}^{(0)}$. It comprises two components.
The first is a semantic critic that verifies whether the composed control sequence remains consistent with the target label, the event order, and the fine-grained instructions.
The second is a set of rule-based physiological checks that validate plausibility with respect to blink duration, inter-blink interval, blink asymmetry, AU co-activation constraints, gaze main sequence, and gaze-head coordination.
The critic outputs structured feedback, and approves the current sequence or sends revision signals back to the composition stage, or further back to the planning stage for event re-decomposition when the current events cannot yield a plausible control sequence.
The refined control sequence is denoted by
  $\hat{C}_{1:T}=Q(C_{1:T}^{(0)};y,E,u,\mathcal{R})$, where $\mathcal{R}$ denotes the set of semantic and  physiological rules, and $Q(\cdot)$ denotes the critic-guided refinement process.
For example, the critic may output ``\textit{semantic consistency: pass; event order: pass; instruction satisfaction: revise gaze\_left and AU7 in Event 2; blink duration: pass; gaze-head coordination: pass}.'' A representative prompt fragment is     
``\textit{You are a behavioral consistency reviewer. Given a target label, staged events, optional fine-grained instructions, and a composed control sequence, check whether the sequence matches the target semantics and satisfies basic physiological constraints. Output the result in a structured format, identify inconsistent AU, gaze, or head-pose channels, and suggest the required revisions.}''                                           
                                                                     
\noindent\textbf{Mapping Layer.}
After the refined control sequence is obtained, the control state at frame $t$ is written as               $\hat{c}_t=[\hat{a}_t,\hat{g}_t,\hat{r}_t]$, where $\hat{a}_t$, $\hat{g}_t$, and $\hat{r}_t$ denote the refined AU, gaze, and head-pose channels, respectively.
Since these channels carry explicit physical meaning, the refined trajectories can be directly projected into facial keypoints. For each frame, AU activations first deform the FLAME mesh linearly, and predefined face indices together with barycentric coordinates are then used to recover the eyelid and eyebrow keypoints. The gaze channels further adjust pupil and iris positions, after which yaw, pitch, and roll are applied to rotate the 3D point set rigidly. The final projection yields 62  keypoints corresponding to the upper and lower eyelids, eyebrows, pupils, and irises, written as      $k_t=\Pi(\hat{a}_t,\hat{g}_t,\hat{r}_t)$.
Mouth keypoints are not controlled explicitly in this stage and are generated in Stage~2 together with audio conditioning.
The final output of Stage~1 is an executable keypoint 
sequence that is consistent with the target label, compatible with the local constraints, and physiologically plausible.  

\subsection{Stage 2: Video Generation with Keypoint Guidance}
\label{sec:stage2}
Stage~2 takes a reference portrait $I_{\mathrm{ref}}$, driving audio $A$, a text prompt $P$, and the facial keypoint sequence $K_{1:T}$, and generates the final portrait animation video $V\in\mathbb{R}^{T\times H\times W\times 3}$.
Following the standard flow-matching sampling process introduced in Sec.~\ref{sec:preliminaries},   $I_{\mathrm{ref}}$ is encoded by the VAE encoder $\mathcal{E}$ into a fixed first-frame latent   $z_{\mathrm{ref}}\in\mathbb{R}^{1\times H'\times W'\times C}$, which occupies the first temporal slot of the initial video latent $x\in\mathbb{R}^{T'\times H'\times W'\times C}$, while the remaining $T'-1$ slots are initialized with Gaussian noise.
The facial keypoints $K$ are first aligned to the reference portrait space by TPS warping and rendered as a pose sequence with the same spatiotemporal resolution as the output video. This pose sequence is then encoded by the VAE encoder $\mathcal{E}$ and a 3D convolutional pose adapter into structural features, which are added to the patch embeddings of the noisy video latents before denoising.
Meanwhile, $z_{\mathrm{ref}}$ is patch-embedded as reference tokens and concatenated with the noisy video tokens.
The text prompt $P$ is encoded by a frozen mT5 text encoder \cite{2020t5} followed by a projection MLP to obtain the text condition $c_{\mathrm{text}}\in\mathbb{R}^{L_s\times d}$, and the driving audio $A$ is encoded by multilingual wav2vec \cite{baevski2020wav2vec} followed by a projection MLP to obtain the audio condition $c_{\mathrm{audio}}\in\mathbb{R}^{L_a\times d}$, where $L_s$ and $L_a$ denote the text-token and audio-token lengths, respectively, and $d$ denotes the condition dimension. During denoising, each DiT block sequentially applies layer normalization, self-attention over video tokens, cross-attention with $c_{\mathrm{text}}$ and $c_{\mathrm{audio}}$ as keys and values, and a feed-forward network, as illustrated in Figure~\ref{fig:pipeline}.
Under these joint conditions, the backbone denoises the video latents to generate the final portrait animation.

\noindent\textbf{Dynamic CFG with eye-region-aware reweighting.} To improve local controllability around the eye region while alleviating global color shift during denoising, we propose a dynamic CFG with eye-region-aware reweighting in both time and space.
Along the temporal dimension, we use a trapezoidal guidance schedule over the denoising progress $\rho=i/N\in[0,1]$, corresponding to the continuous timestep $t$ in the flow-matching formulation, where $N$ is the total number of inference steps and $i$ is the current step index: the guidance weight stays at the higher level $\omega_{\mathrm{hi}}$ when $\rho<\alpha$, decreases linearly from $\omega_{\mathrm{hi}}$ to the lower level $\omega_{\mathrm{lo}}$ when $\alpha\le\rho<\gamma$, and remains at $\omega_{\mathrm{lo}}$ thereafter, where $\alpha$ and $\gamma$ define the transition interval.
This schedule applies stronger guidance during early denoising steps, where low-frequency components such as global color emerge, to fix the overall appearance early, and subsequently reduces the guidance to avoid the color shift caused by a uniformly large CFG scale.

Along the spatial dimension, we assign stronger guidance to the eye region and weaker guidance to the background.
Let $c_{\mathrm{eye}}$ denote the midpoint of the two iris centers extracted from the facial
keypoints $K$ and let $d_{\mathrm{eye}}$ denote the inter-eye distance, both projected to the latent coordinate space of the VAE encoder $\mathcal{E}$. We define a Gaussian spatial weighting map over latent positions $(x,y)$ as:                                    \begin{equation}                               G_f(x,y)=\exp\left(-\frac{\|(x,y)-c_{\mathrm{eye}}\|^2}{2(\kappa d_{\mathrm{eye}})^2}\right),                   
  \end{equation}  where $\kappa$ controls the Gaussian width.
Let $\omega_{\mathrm{bg}}$ denote the base CFG weight in the background.
The final guidance scale at denoising progress $\rho$ and spatial position $(x,y)$ is:        \begin{equation}                               \Omega_f(x,y,\rho)=\omega_t(\rho)\,G_f(x,y)+\omega_{\mathrm{bg}}\bigl(1-G_f(x,y)\bigr).       \end{equation}                                 During inference, the conditional and unconditional noise predictions are fused with $\Omega_f$ as the spatially varying guidance weight.
This design concentrates stronger guidance on the eye region while preserving the     
background at the base weight, improving local controllability of gaze, blink, and eyebrow motion without       
degrading global visual quality.

\noindent\textbf{KTO-based Refinement for Boundary Cases.}
Even with the dynamic CFG and eye-region-aware reweighting described above, the model still occasionally struggles with boundary cases, including imprecise asymmetric eyebrow control and degraded identity consistency under large-angle head movements. Similar failures also appear in other challenging long-tail patterns, such as rapid, irregular gaze movements and blinks.
To alleviate these issues, we further refine the DiT generation backbone with KTO. 
Specifically, we curate a KTO training set $\mathcal{D}_{\mathrm{KTO}}=\{(c_i,y_i,s_i)\}_{i=1}^{M}$ for these   
boundary cases, where $c_i$ denotes the conditional input (reference portrait, audio, keypoints, and text prompt), $y_i$ denotes the generated video latent, and $s_i\in\{0,1\}$ indicates whether the sample is desirable or undesirable.
Desirable samples are real videos collected from boundary cases, whereas undesirable
samples are generated by the current model under the same conditional inputs and selected when they exhibit imprecise eye-region or head-motion control, or degraded identity consistency during large-angle head movement. Following the KTO objective defined in Sec.~\ref{sec:preliminaries}, the implicit reward is $r_\theta(c_i, y_i) = \log
\frac{\pi_\theta(y_i|c_i)}{\pi_{\mathrm{ref}}(y_i|c_i)}$, where $\pi_\theta$ is the flow-matching velocity network parameterized by the DiT backbone and pose encoder, and $\pi_{\mathrm{ref}}$ is the frozen checkpoint
before KTO refinement. Desirability is determined by control accuracy and identity consistency.

\begin{figure*}[!t]
    \centering
\includegraphics[width=0.90\linewidth]{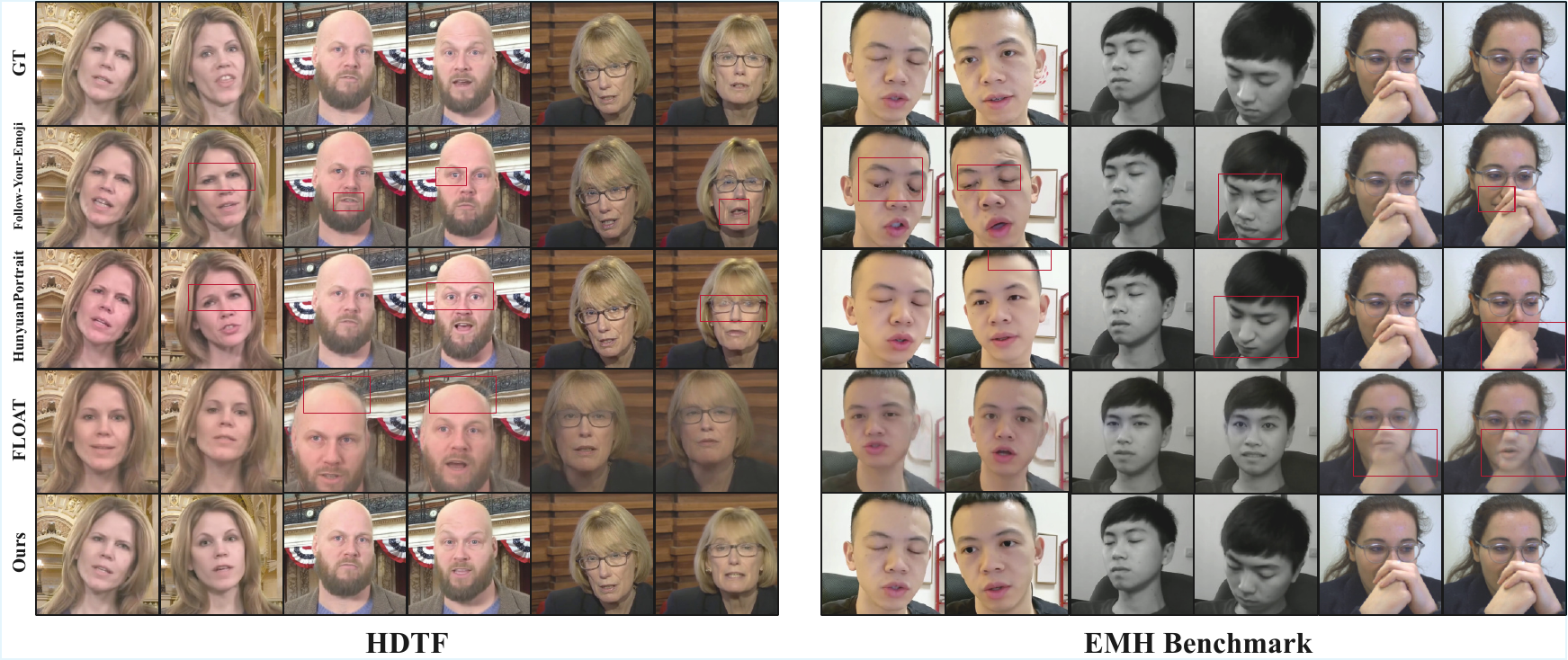}
\caption{Qualitative comparison on HDTF (left) and the EMH benchmark (right) against representative baselines. Red boxes highlight typical failure cases of the baselines, such as crown artifacts and imprecise eyebrow control. Zoom in for more details.}
    \label{fig:qualitative}
\end{figure*}

\section{Experiments}
\subsection{Experiment Setup}
\label{sec:data}
\noindent\textbf{Implementation Details.}
All experiments were conducted on 8 NVIDIA H200 GPUs. The DiT backbone was initialized from Wan2.2~\cite{wan2025wan} and fine-tuned with the pose encoder using AdamW at $1\times10^{-5}$ for 7 GPU days. For KTO refinement, the learning rate was $5\times10^{-8}$ with $\beta=625$, combining the KTO and flow-matching losses at equal weight, for 2 GPU days. Inference used 40 denoising steps with $\omega_{\mathrm{hi}}=8.0$,      
  $\omega_{\mathrm{lo}}=4.0$, $\alpha=0.25$, $\gamma=0.55$, $\kappa=1.0$, and $\omega_{\mathrm{bg}}=1.0$. The CoT agents used Gemini 3.0 in JSON mode.
  
\noindent\textbf{Datasets.}
For training DiT, we used TalkVid~\cite{chen2025talkvidlargescalediversifieddataset} and DH-FaceVid-1K~\cite{di2024facevid}.
For KTO-based refinement, we curated a preference dataset covering four boundary-case categories: large irregular head motions (400 clips of 3--5\,s from DH-FaceDrasMvVid \cite{sun2024uniavatar} with yaw $>40^\circ$, pitch $>30^\circ$, or roll $>30^\circ$), and three self-collected categories from 15 volunteers covering asymmetric eyebrow motion, erratic gaze drift, and rapid blinking. Each category contained 400 clips (350 train / 50 test)  
  with no identity overlap.
For evaluation, HDTF~\cite{zhang2021flow} measured standard visual quality and lip synchronization, and the EMH benchmark
measured fine-grained eye-region and head-motion pattern generation across 6 core emotions and 6 beyond-emotion categories. The training and test identities were non-overlapping.

\begin{table}[!t]
    \centering
    \footnotesize
    \setlength{\tabcolsep}{3.5pt}
     \caption{Quantitative comparison with existing methods on HDTF. We reported FID, FVD, and LPIPS for visual quality, Sync-C for lip synchronization, ID-Sim for identity consistency, and Eye-LMD for     
  eye-region control accuracy. Best results are in \textbf{bold} and second-best are \underline{underlined}.}  
    \label{tab:hdtf}
    \resizebox{\columnwidth}{!}{
    \begin{tabular}{lcccccc}
        \toprule
        Method & FID $\downarrow$ & FVD $\downarrow$ & LPIPS $\downarrow$ & Sync-C $\uparrow$ & ID-Sim $\uparrow$ & Eye-LMD $\downarrow$ \\
        \midrule
        EDTalk \cite{tan2025edtalk} & 98.77 & 142.88 & 0.1749 & 5.67 & 0.8336 & 0.0210 \\
        Follow-Your-Emoji \cite{ma2024follow} & \underline{18.22} & 72.11 & 0.0929 & 4.83 & 0.8609 & \underline{0.0173} \\
        Sonic \cite{ji2025sonic} & 24.66 & 51.65 & 0.1075 & \textbf{7.43} & 0.8609 & 0.0363 \\
        Hallo3 \cite{cui2024hallo3} & 53.85 & 45.76 & 0.1885 & 5.92 & 0.8622 & 0.0255 \\
        HunyuanPortrait \cite{xu2025hunyuanportrait} & 18.68 & 39.79 & 0.0962 & 6.22 & 0.8837 & 0.0293 \\
        FantasyTalking \cite{wang2025fantasytalking} & 44.79 & 51.46 & \underline{0.0887}& 6.95 & \underline{0.8957} & 0.0346 \\
        FLOAT \cite{ki2025float} & 72.77 & 99.25 & 0.2676 & 6.92 & 0.7651 & 0.0781 \\
        ACTalker \cite{hong2025audio} & 27.92 & \underline{37.64 } & 0.1934 & 7.02 & 0.8651 & 0.0217 \\
        \rowcolor{blue!10}
        \textbf{Ours} & \textbf{16.68} & \textbf{32.90} & \textbf{0.0633} & \underline{7.15} & \textbf{0.9214} & \textbf{0.0107} \\
        \bottomrule
    \end{tabular}}
\end{table}

\noindent\textbf{Data Pre-processing.}
A unified pre-processing pipeline was applied to all training and test data. Each video was first converted to 25 FPS, and the corresponding audio was resampled to 16 kHz. The facial region was then cropped and resized to a resolution of $512\times512$. Training and evaluation were performed after this pre-processing.

\noindent\textbf{Baselines.}
We compared against eight representative methods spanning three input categories: audio-driven (Sonic~\cite{ji2025sonic},  FLOAT~\cite{ki2025float}, Hallo3~\cite{cui2024hallo3}, FantasyTalking~\cite{wang2025fantasytalking}), label-driven  (EDTalk~\cite{tan2025edtalk}), and video-driven (Follow-Your-Emoji~\cite{ma2024follow}, HunyuanPortrait~\cite{xu2025hunyuanportrait}, and ACTalker~\cite{hong2025audio}).
All baselines were evaluated using their     
official implementations and pre-trained checkpoints.

\noindent\textbf{Evaluation Metrics.}
For HDTF \cite{zhang2021flow}, we used FID \cite{heusel2017gans}, FVD \cite{wang2018video}, and LPIPS \cite{zhang2018unreasonable} to evaluate visual quality, SyncNet Confidence \cite{li2024latentsync} (Sync-C) to evaluate audio-visual synchronization, ArcFace cosine similarity \cite{deng2019arcface} (ID-Sim) to evaluate identity preservation, and eye-region landmark distance (Eye-LMD) \cite{lugaresi2019mediapipe} to evaluate eye-region controllability. 
For the EMH benchmark, we reported BRISQUE \cite{6272356} for no-reference visual quality, DSL-FIQA \cite{chen2024dsl} for face quality, ID-Sim, Eye-LMD, and two AU-level metrics: AU-F1 for activation correctness and AU-Temp for temporal trajectory fidelity. Let $\hat{A}_u(t)$ and $A_u(t)$ denote the intensity trajectories of the $u$-th AU extracted from the generated and ground-truth videos, respectively, and let $S_y$ be the target AU set for pattern category $y$. We define binary activations $a_u=\mathbb{I}(\max_t A_u(t)>\tau_u)$ and $\hat{a}_u=\mathbb{I}(\max_t \hat{A}_u(t)>\tau_u)$ with per-AU threshold $\tau_u$. AU-F1 is then the harmonic mean of precision and recall over the $U$ AUs:
\begin{equation}
\textstyle
P=\frac{\sum_u\mathbb{I}(\hat{a}_u{=}1 \wedge a_u{=}1)}{\sum_u\mathbb{I}(\hat{a}_u{=}1)},\;
R=\frac{\sum_u\mathbb{I}(\hat{a}_u{=}1 \wedge a_u{=}1)}{\sum_u\mathbb{I}(a_u{=}1)},\;
\mathrm{AU\mbox{-}F1}=\frac{2PR}{P+R},
\end{equation}
which measures whether the correct AU set is activated. For temporal AU alignment, we define
\begin{equation}
\textstyle
\mathrm{AU\mbox{-}Temp}=1-\frac{1}{|S_y|}\sum_{u\in S_y}\mathrm{DTW}(\hat{A}_u,A_u)\,/\,Z_u,
\end{equation}
where $\mathrm{DTW}(\cdot)$ is the dynamic time warping distance and $Z_u$ is a normalization constant.

\subsection{Quantitative Comparison}
\label{sec:quantitative}
\noindent\textbf{Performance on HDTF Dataset.}
As shown in Table~\ref{tab:hdtf}, our method achieved the best FID (16.68), FVD (32.90), LPIPS (0.0633), ID-Sim (0.9214), and Eye-LMD (0.0107) among all methods. The Sync-C score (7.15) was comparable to Sonic (7.43). The improvement in Eye-LMD was especially notable, with a 38\% relative reduction over the second-best Follow-Your-Emoji (0.0173), confirming that our framework delivers more precise eye-region control while maintaining competitive visual quality and lip synchronization on the standard portrait animation benchmark \cite{zhang2021flow}.

\noindent\textbf{Performance on the EMH Benchmark.}             As shown in Table~\ref{tab:proposed_benchmark}, we evaluated all methods on 12 categories of the EMH benchmark. Video-driven methods such as HunyuanPortrait and Follow-Your-Emoji outperformed audio- and label-driven methods such as Sonic, FLOAT, and EDTalk in AU-F1 and AU-Temp, since these two metrics assess the consistency of gaze, eyelid, blink, and eyebrow trends, which audio- and label-driven methods cannot explicitly control. When driven by ground-truth facial keypoints, our DiT also outperformed HunyuanPortrait in AU-F1 (0.9303 vs. 0.9145) and AU-Temp (0.7830 vs. 0.7475), because the keypoint input directly encodes fine-grained gaze, eyelid, blink, and eyebrow signals. The inverse pipeline caused only a mild AU-F1 drop (0.9192 vs. 0.9303). Using keypoints generated entirely by Stage~1, Ours (Full) still reached 0.9017 AU-F1 and 0.7397 AU-Temp, comparable to HunyuanPortrait and substantially above all audio- and label-driven methods, confirming that the agent-generated keypoints can produce accurate fine-grained eye-region dynamics. In addition, our method achieved better face image quality (BRISQUE 34.20 vs. 46.33 of FLOAT), higher identity consistency (ID-Sim 0.9129 vs. 0.9067 of FantasyTalking), and lower Eye-LMD (0.0145 vs. 0.0192 of HunyuanPortrait).


\begin{table}[t]
    \centering
    \footnotesize
    \setlength{\tabcolsep}{3.0pt}
    \caption{Performance comparison on the EMH benchmark. $\downarrow$: lower is better; $\uparrow$: higher is better. \textbf{Ours (GT)} uses GT keypoints directly; \textbf{Ours (Inverted)} constructs the prototype library via AU inversion from GT; \textbf{Ours (Full)} is the complete pipeline with Stage~1 keypoint generation.}
    \label{tab:proposed_benchmark}
    \resizebox{\columnwidth}{!}{
    \begin{tabular}{lcccccc}
        \toprule 
        Method & BRISQUE $\downarrow$ & DSL-FIQA $\uparrow$ & ID-Sim $\uparrow$ & Eye-LMD $\downarrow$ & AU-F1 $\uparrow$ & AU-Temp $\uparrow$ \\
        \midrule
        EDTalk \cite{tan2025edtalk} & 49.74 & 0.3838 & 0.8601 & 0.0313 & 0.5449 & 0.5255 \\
        Follow-Your-Emoji \cite{ma2024follow} & 47.52 & 0.4144 & 0.8843 & 0.0237 & 0.8816 & 0.6936 \\
        Sonic \cite{ji2025sonic} & 52.59 & 0.4761 & 0.9044 & 0.0395 & 0.5707 & 0.5249 \\
        Hallo3 \cite{cui2024hallo3} & 50.15 & 0.4625 & 0.8964 & 0.0393 & 0.5487 & 0.4566 \\
        HunyuanPortrait \cite{xu2025hunyuanportrait} & 49.80 & 0.4859 & 0.9019 & 0.0192 & 0.9145 & 0.7475 \\
        FantasyTalking \cite{wang2025fantasytalking} & 53.21 & 0.4598 & 0.9067 & 0.0305 & 0.5554 & 0.4758 \\
        FLOAT \cite{ki2025float} & 46.33 & 0.4710 & 0.8580 & 0.0746 & 0.5297 & 0.3251 \\
        ACTalker \cite{hong2025audio} & 61.17 & 0.4686 & 0.7308 & 0.0534 & 0.5212 & 0.3188 \\
        \midrule
        \rowcolor{blue!10}
        Ours (GT) & \underline{32.84} & \underline{0.5192} & \textbf{0.9362} & \textbf{0.0112} & \textbf{0.9303} & \textbf{0.7830} \\
        \rowcolor{blue!10}
        Ours (Inverted) & \textbf{32.73} & \textbf{0.5266} & \underline{0.9217} & \underline{0.0133} & \underline{0.9192} & \underline{0.7791} \\
        \rowcolor{blue!10}
        Ours (Full) & 34.20 & 0.5184 & 0.9129 & 0.0145 & 0.9017 & 0.7397 \\
        \bottomrule
    \end{tabular}}
\end{table}

\begin{figure}[!t]
    \centering
    \includegraphics[width=0.80\linewidth]{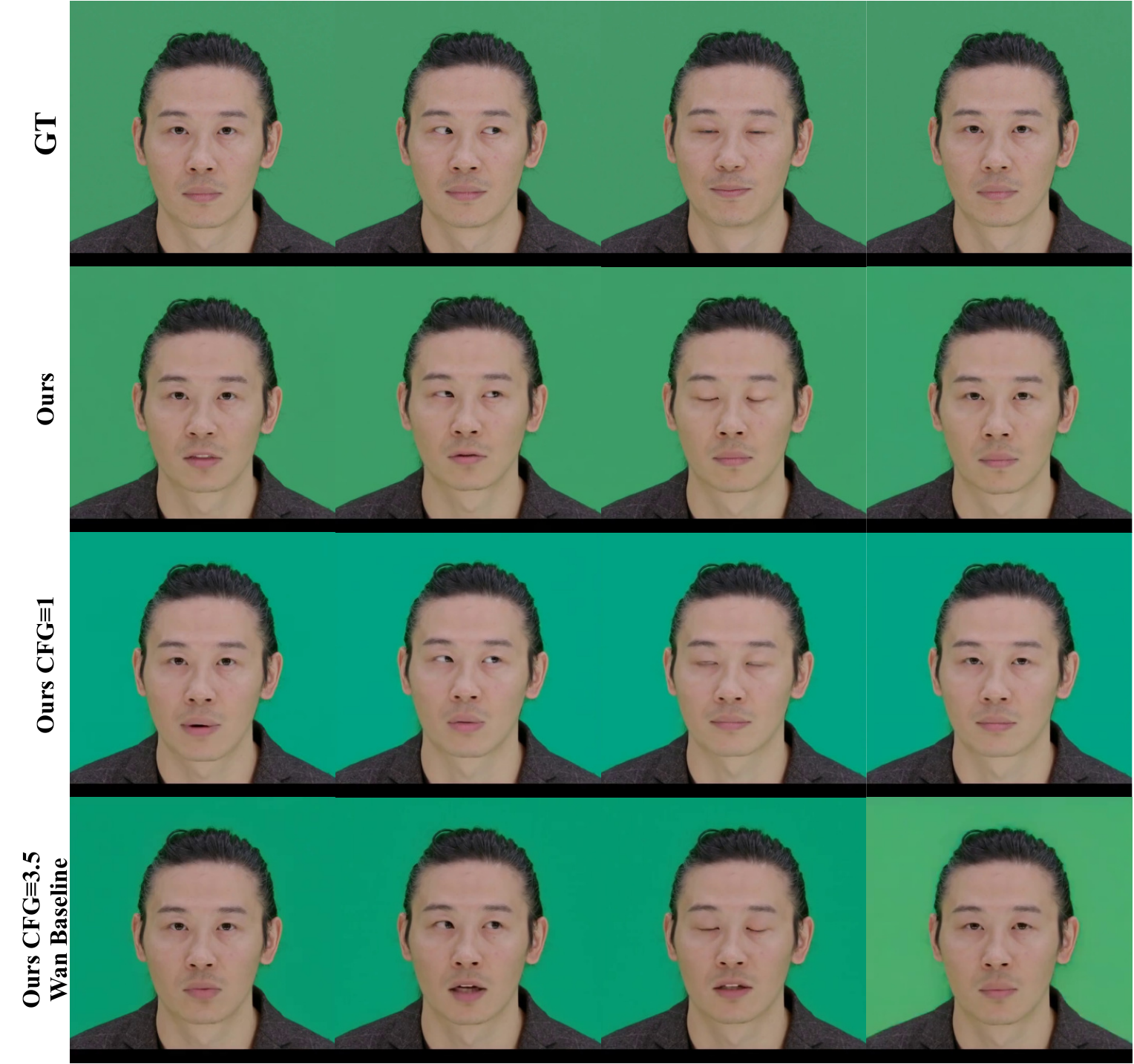}
    \caption{Ablation study for CFG strategies. We selected images with pure green backgrounds as test cases to expose the global color shift issue, where sharing a single CFG scale across the entire image often leads to undesired color deviation in both the background and facial regions.}
    \label{fig:color}
\end{figure}

\begin{figure}[!t]
    \centering
    \includegraphics[width=0.85\linewidth]{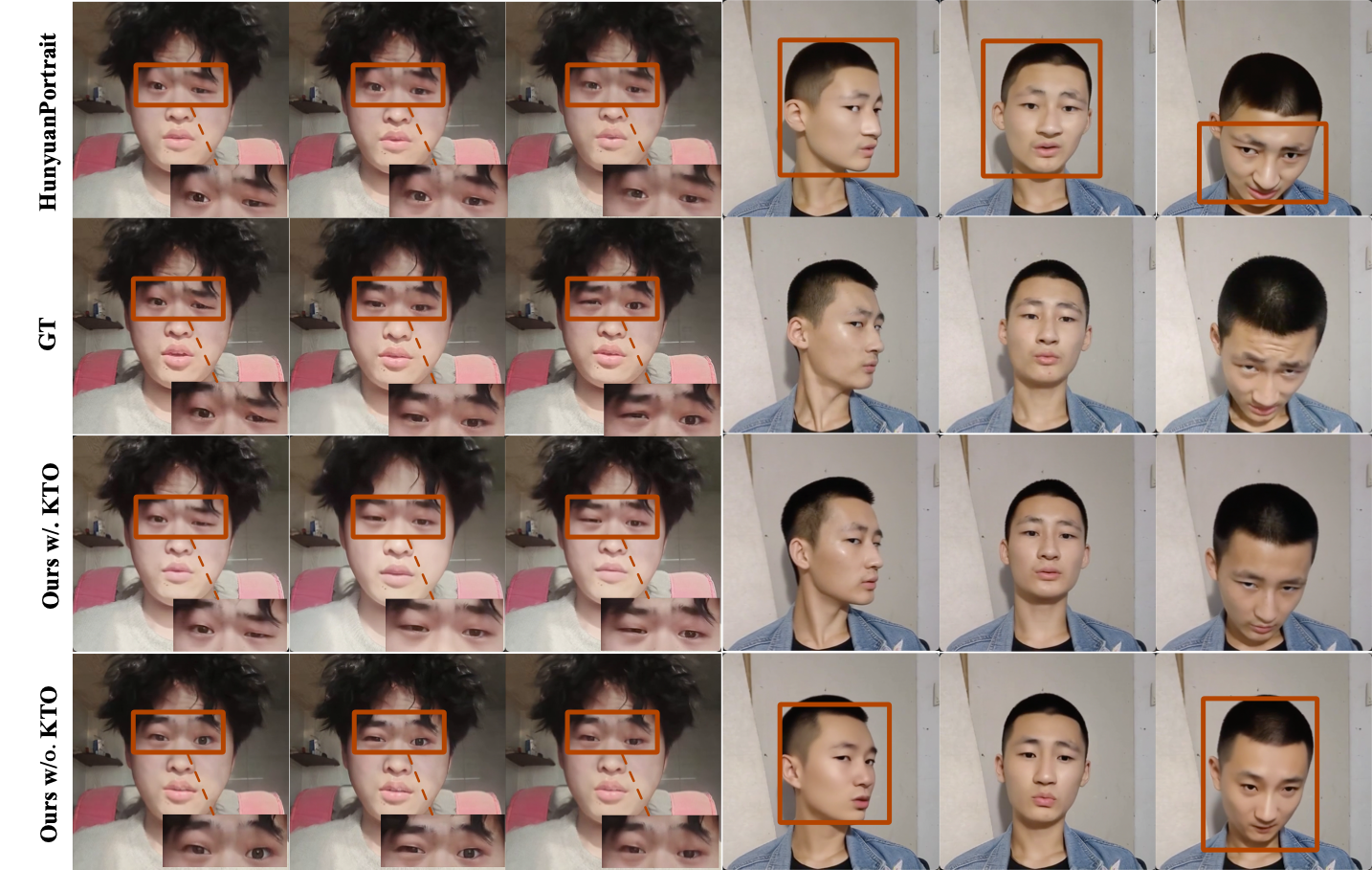}
    \caption{Ablation study for the KTO refinement on the controllability of eye-region irregular motions (\ie, asymmetric eyebrow raising) and the identity consistency under large-angle head movements. Red boxes highlight the key local regions for comparison. Please zoom in for more details.}
    \label{fig:kto}
\end{figure}

\subsection{Qualitative Comparison}
\noindent\textbf{Comparison on HDTF.}                                                            We presented qualitative comparisons on HDTF in Figure~\ref{fig:qualitative}~(left). FLOAT was label-driven and lacked fine-grained eye-region control.
Follow-Your-Emoji exhibited noticeable skin and eye color drift due to its globally shared CFG strategy.
HunyuanPortrait occasionally failed to control eyebrows precisely because its training objective targets full-face control rather than fine-grained eye-region control.
Our method maintained superior identity consistency and more precise eye-region control.                       

\noindent\textbf{Comparison on the EMH Benchmark.}                                            We further compared methods on the EMH benchmark across social engagement, drowsiness, and cognitive effort (Figure~\ref{fig:qualitative}, right).
FLOAT produced visible background distortion and mouth artifacts under beyond-emotion states. Follow-Your-Emoji generated eye-region artifacts, such as excessive eyelid pulling, and suffered from identity drift during drowsy head nodding.             
HunyuanPortrait also exhibited identity drift during large head motions and edge artifacts. Our method consistently preserved identity while generating accurate eye-region dynamics across all three categories.
We provide qualitative comparisons for the remaining EMH categories in the \textit{supp}.

\subsection{Ablation Studies}
\noindent\textbf{Ablation on Stage~1 Components}                                                As shown in Table~\ref{tab:ablation}, directly mapping labels to keypoints via a prediction transformer produced reasonable static poses but lacked temporal variation, leading to the lowest AU-Temp (0.5676).  Rule-based retrieval from the prototype library improved temporal dynamics but occasionally combined physiologically incompatible prototypes (e.g., raised eyebrows with droopy eyelids), because it selects within a single category without modeling individual variation. Adding the planning agent provided structured temporal decomposition, and the composition agent further improved both AU-F1 and AU-Temp by retrieving category-specific prototypes. The critic agent eliminated occasional hallucinations from the composition agent, yielding the best overall performance (AU-F1 0.9017, Eye-LMD 0.0145).

\noindent\textbf{Effect of Dynamic CFG Strategy.}                   As shown in Figure~\ref{fig:color}, when the reference portrait has a uniform background color, baseline CFG schedules with uniform scales (1.0, the Wan2.2 default 3.5, and 5.0) caused noticeable global color shifts that affected both the facial region and background. In contrast, our dynamic CFG with eye-region-aware reweighting alleviated this issue and maintained color consistency in both regions. As reported in Table~\ref{tab:ablation}, the temporal schedule improved BRISQUE from 34.73 to 34.51 and ID-Sim from 0.8876 to 0.9073, confirming that fixing the overall appearance in early denoising steps reduces the color drift. The spatial eye-region weighting reduced Eye-LMD from 0.0207 to 0.0158 and raised AU-F1 from 0.8823 to 0.8983, confirming that concentrating guidance on the eye region improves gaze, blink, and eyebrow controllability. Both components complement each other, and their combination achieved the best results across all metrics.

\begin{table}[!t]
    \centering
    \footnotesize
    \setlength{\tabcolsep}{3.0pt}
    \caption{Ablation study on the proposed EMH benchmark.}
    \label{tab:ablation}
    \resizebox{\columnwidth}{!}{
    \begin{tabular}{lcccccc}
        \toprule
        Variant & BRISQUE $\downarrow$ & DSL-FIQA $\uparrow$ & ID-Sim $\uparrow$ & Eye-LMD $\downarrow$ & AU-F1 $\uparrow$ & AU-Temp $\uparrow$ \\
        \midrule
        \multicolumn{7}{l}{\textit{Stage-1 ablation}} \\
        \rowcolor{gray!10}
        Direct label-to-keypoints & 40.25 & 0.4092 & 0.8803 & 0.0253 & 0.8535 & 0.5676 \\
        \rowcolor{gray!10}
        Rule-based retrieval & 39.31 & 0.4488 & 0.8817 & 0.0224 & 0.8812 & 0.5899 \\
        \rowcolor{gray!10}
        Agent 1 only & 36.78 & 0.4785 & 0.8834 & 0.0195 & 0.8346 & 0.7112 \\
        \rowcolor{gray!10}
        Agent 1 + Agent 2 w/o Critic & 34.83 & 0.5081 & 0.8932 & 0.0173 & 0.8821 & 0.7135 \\
        \midrule
        \multicolumn{7}{l}{\textit{Stage-2 CFG ablation}} \\
        \rowcolor{gray!10}
        Fixed CFG ($s{=}3.5$) & 34.73 & 0.5062 & 0.8876 & 0.0207 & 0.8823 & 0.7275 \\
        \rowcolor{gray!10}
        Temporal reweighting only & 34.51 & 0.5078 & 0.9073 & 0.0193 & 0.8835 & 0.7301 \\
        \rowcolor{gray!10}
        Spatial reweighting only & 34.35 & 0.5139 & 0.8945 & 0.0158 & 0.8903 & 0.7368 \\
        \midrule
        \rowcolor{blue!10}
        \textbf{Full pipeline} & \textbf{34.20} & \textbf{0.5184} & \textbf{0.9129} & \textbf{0.0145} & \textbf{0.9017} & \textbf{0.7397} \\
        \bottomrule
    \end{tabular}}
\end{table}


\noindent\textbf{Impact of KTO Refinement.}                     As shown in Figure~\ref{fig:kto}, we selected two representative boundary cases for qualitative comparison: asymmetric eyebrow motion and near-profile head turning. For asymmetric eyebrow raising, HunyuanPortrait failed to produce the intended asymmetry, Ours w/o KTO was slightly better but still imprecise, and Ours w/ KTO clearly achieved the target asymmetric pattern.
For near-profile head poses, HunyuanPortrait exhibited noticeable identity drift, Ours w/o KTO alleviated this drift, and Ours w/ KTO produced results closer to the ground truth. As reported in Table~\ref{tab:kto_comparison}, KTO refinement reduced Eye-LMD from 0.0377 to 0.0311 and increased ID-Sim from 0.9089 to 0.9394, confirming improved control accuracy and identity consistency on boundary cases. 

\begin{table}[!t]
    \centering
    \footnotesize
    \setlength{\tabcolsep}{3.0pt}
    \caption{Comparison on the boundary-case subset of the EMH benchmark.}
    \label{tab:kto_comparison}
    \resizebox{\columnwidth}{!}{
    \begin{tabular}{lcccccc}
        \toprule
        Method & BRISQUE $\downarrow$ & DSL-FIQA $\uparrow$ & ID-Sim $\uparrow$ & Eye-LMD $\downarrow$ & AU-F1 $\uparrow$ & AU-Temp $\uparrow$ \\
        \midrule
        ACTalker \cite{hong2025audio} & 52.52 & 0.4701 & 0.8716 & 0.1046 & 0.5122 & 0.3478 \\
        HunyuanPortrait \cite{xu2025hunyuanportrait} & 37.46 & 0.4955 & 0.9017 & 0.0395 & 0.8332 & 0.6637 \\
        Follow-Your-Emoji \cite{ma2024follow} & 44.43 & 0.4309 & 0.9048 & 0.0407 & 0.8454 & 0.7003 \\
        \rowcolor{gray!10}
        Ours w/o KTO & 34.27 & 0.5308 & 0.9089 & 0.0377 & 0.8465 & 0.7235 \\
        \rowcolor{blue!10}
        \textbf{Ours w/ KTO} & \textbf{30.95} & \textbf{0.5574} & \textbf{0.9394} & \textbf{0.0311} & \textbf{0.9121} & \textbf{0.8238} \\
        \bottomrule
    \end{tabular}}
\end{table}

\subsection{User Study}
\label{sec:user_study}
We conducted a user study comparing our method with FLOAT, Hallo3, and ACTalker on motion realism, motion diversity, identity consistency, and overall preference (see \textit{supp}).

\section{Conclusion}
We presented CogPortrait, a two-stage framework for fine-grained eye-region control in portrait animation. Stage~1 compiles high-level labels into facial keypoints via prototype retrieval; Stage~2 generates the animation conditioned on these keypoints. We further introduce dynamic CFG and KTO-based refinement to improve generation quality on long-tail cases. Beyond core emotions, CogPortrait also supports cognitive states such as cognitive effort and drowsiness.


\clearpage


\bibliographystyle{ACM-Reference-Format}
\bibliography{sample-base}


\end{document}